\documentclass[conference]{IEEEtran}
\IEEEoverridecommandlockouts
\usepackage{cite}
\usepackage{amsmath,amssymb,amsfonts}
\usepackage{algorithmic}
\usepackage{multirow}
\usepackage{graphicx}
\usepackage{textcomp}
\usepackage{url}
\usepackage{listings}
\usepackage{tikz}
\usetikzlibrary{shapes}
\usepackage{xcolor}
\def\BibTeX{{\rm B\kern-.05em{\sc i\kern-.025em b}\kern-.08em
    T\kern-.1667em\lower.7ex\hbox{E}\kern-.125emX}}
    
\usepackage{eurosym}
\usepackage{booktabs}
\usepackage{tikz}
\newcounter{row}
\newcounter{col}
\newcommand\setrow[4]{
  \setcounter{col}{1}
  \foreach \n in {#1, #2, #3, #4} {
    \edef\x{\value{col} - 0.5}
    \edef\y{4.5 - \value{row}}
    \node[anchor=center] at (\x, \y) {\n};
    \stepcounter{col}
  }
  \stepcounter{row}
}    
    
\usepackage{listingsutf8}

\lstdefinelanguage{xcsp}{
  keywords={region,literals,domain,annotation,extension,intension, tuples, fuzzy,relation,interval,point,agents,agent,allDifferent,count,ordered,vars,ctrs,comm,nodes,edges,arcs,minimize,maximize,linear,operator,allEqual,among,atLeast,atMost,exactly,limit,cumulative,circuit,regular,mdd,hamming,channel,element,values,transitions,durations,indexes,permutation,nValues,clause,cube,instantiation,sort,lex,precedence,cardPath,slide,slidingAmong,slidingSum,sequence,gsc,seqbin,binPacking,minimum, hardTemplate,softTemplate,capacities,loads,intervals,costMatrix,knapsack,allDisjoint,overlap,aggregate,dbd,output,max,valHeuristic,lastConflict,vals,BC, AC, FC, filtering, width, widths, stretch,start,final,balance,path,tree, root, succs, image, allDistant, roots, capacity, bins, sizes, and, or, except, matrix, sumCosts, increasing, lists,terminal,rules,grammar,mapping,cards, ranges, rowCards, colCards, indexes, not, allIntersecting, costMeasure, accept, limits, relation, sum, comparison, operand, conditions, nCircuits, nPaths, nTrees, rowOccurs, colOccurs, occurs, noOverlap,total, number, cardinality, static, ifThen, ifThenElse, smart, row, equal, lessThan, constraints, list, variables, instance, args
},
  keywordstyle=\color{black}\bfseries,
  ndkeywords={prepro,search,restarts,varHeuristic,decision, stochastic,forall,exists,array,block,group,lc,combination,defaultDegree,startIndex, reifiedBy, hreifiedFrom, hreifiedTo, as, measure, degree, threshold, cutoff, varsModel, ctrsModel, commModel, for, case, closed, rank, restriction, circular, offset, collect, window, violable, order , defaultCost, solution, optimum, violationCost, violationMeasure},
  ndkeywordstyle=\color{black}\bfseries,
  identifierstyle=\color{black},
  sensitive=false,
  comment=[l]{//},
  morecomment=[s]{<!--}{-->},
  commentstyle=\color{black}\ttfamily,
  stringstyle=\color{black}\ttfamily,
  morestring=[b]',
  morestring=[b]",
  escapechar=@,
  showstringspaces=false,
  xleftmargin=8pt,xrightmargin=0pt,
  breaklines=true,basicstyle=\ttfamily,inputencoding=utf8/latin9,texcl
}

\lstdefinelanguage{json}{
    basewidth  = {.6em,0.6em},
    basicstyle=\normalfont\ttfamily,
    breaklines=true,
    morestring=[b]',
    morestring=[b]", 
    sensitive=false,
    stringstyle=\color{dgreen}\ttfamily, 
    escapechar=!,
    showstringspaces=false,
    xleftmargin=8pt,xrightmargin=1pt,
    breaklines=true,basicstyle=\ttfamily\small,inputencoding=utf8/latin9,texcl
}

\lstnewenvironment{xcsp}{\lstset{language=xcsp}}{} 
\lstnewenvironment{json}{\lstset{language=json}}{}

\begin{document}

\title{Ludii and XCSP: \\ Playing and Solving Logic Puzzles
}

\author{
\IEEEauthorblockN{C\'{e}dric Piette \\~\\}
\IEEEauthorblockA{\textit{Centre de Recherche en Informatique de Lens} \\
\textit{Universit\'{e} d'Artois}\\
Lens, France \\
\texttt{piette@cril.fr}}
\and
\IEEEauthorblockN{{\'E}ric Piette, Matthew Stephenson, Dennis J.N.J. Soemers, \\ Cameron Browne}
\IEEEauthorblockA{\textit{Department of Data Science and Knowledge Engineering} \\
\textit{Maastricht University}\\
Maastricht, the Netherlands \\
\texttt{\{eric.piette,matthew.stephenson,dennis.soemers,}\\\texttt{ cameron.browne\}@maastrichtuniversity.nl}}
}

\maketitle

\begin{abstract}
Many of the famous single-player games, commonly called puzzles, can be shown to be NP-Complete. Indeed, this class of complexity contains hundreds of puzzles, since people particularly appreciate completing an intractable puzzle, such as Sudoku, but also enjoy the ability to check their solution easily once it's done. For this reason, using constraint programming is naturally suited to solve them. 
In this paper, we focus on logic puzzles described in the Ludii general game system and we propose using the XCSP formalism in order to solve them with any CSP solver.
\end{abstract}


\section{Introduction}
Since the Nikoli company\footnote{Nikoli: \url{nikoli.co.jp/en/puzzles/}} publishes many puzzles in different newspapers, such as the well-known Sudoku, solving pure deduction puzzles (mainly Japanese logic puzzles \cite{Japanese06}) became a widespread pastime around the world.

In the context of General Game Playing (GGP), where artificial agents have to be capable of playing a wide variety of games (including puzzles) \cite{pitrat68}, Monte Carlo Tree Search (MCTS) \cite{Browne2012} is now considered as one of the best approaches in the absence of domain specific knowledge \cite{finnsson10}. For puzzles, two main previous works on MCTS exist. The first one, called Single Player MCTS used in the \textit{SameGame} player has been introduced in \cite{Schadd12}. This variant adds a new term in the UCB formula, and uses a heuristically guided default policy for simulations in order to optimize their resolution. The second work defines an adapted UCT algorithm for the combinatorial optimisation problem of feature selection \cite{Gaudel10}. 

However, these two variants -- and MCTS in general -- do not tend to perform well on pure deduction puzzles. %, mainly because 
The nature of these puzzles \cite{Browne15} may not be suitable for this paradigm, for instance due to most of them exhibiting only a single solution.

Regarding complexity, many of the (logic) puzzles, are shown to be NP-complete \cite{costa18}. In this context, constraint programming (CP) appears to be a particularly effective means of solving them \cite{Sullivan07, Celik09}.
%, as reflected by some of the previous work on specific logic puzzles, e.g. \cite{Sullivan07} or \cite{Celik09}. 
Moreover, CP is well-studied, leads to highly explainable solutions, and finding solutions is typically efficient \cite{Jaffar94}. 
In this paper, we present how the General Game system Ludii \cite{Piette19} can model and play any logic puzzle through the XCSP formalism \cite{BoussemartLP16}, enabling the use of any compatible state-of-the-art solver to efficiently solve them.

\section{Ludii and XCSP}

In this section, we briefly describe Ludii and XCSP, as well as the translation process from Ludii to XCSP for puzzles.

\subsection{Ludii}
Within the context of the Digital Ludeme Project\footnote{Digital Ludeme Project: \url{http://www.ludeme.eu/}} \cite{Browne2018ModernTechniques}, a new general game system called Ludii was recently proposed. This system is based on a ludemic modelisation and a class grammar approach for games \cite{BrowneB16}.

In Ludii, a game is given by a 3-tuple of ludemes $\mathcal{G} = \langle \mathit{Mode, Equipment, Rules} \rangle$. $Mode$ denotes a finite set of $k$ deterministic players. $\mathit{Equipment} = \langle C^t, C^p \rangle$ denotes a set of containers $C^t$, and a set of components $C^p$. Finally, $Rules$ defines the operations of the game, which is split in three distinct parts: $\mathit{start}$, $\mathit{play}$, and $\mathit{end}$.

In contrast to the standard General Game system using the Game Description Language (GDL) \cite{love08}, the ludemic approach offers the capability to model puzzle rules. Each ludeme encapsulates the concepts commonly used like arithmetic operators, inequality between many grid cells or specific regions (row, column, etc). An example Ludii description for a Sudoku puzzle on a $4$$\times$$4$ grid is provided in Figure \ref{fig:ludiiSudoku}, with the puzzle itself depicted in Figure \ref{fig:Sudoku}.

Ludii provides many benefits relative to GDL. Among them, the description of the games are self-explanatory to non-specialist readers and all games available in the system can be played by both humans and/or AI. Moreover, even if Ludii uses MCTS as the core method for AI move planning, the structural composition of the games, and more particularly puzzles, can be exploited to make use of different paradigms such as CP.

\begin{figure}[!t]
\scriptsize
\lstset{numbers=left, numberstyle=\tiny,stepnumber=1,numberfirstline=false, numbersep=5pt,basicstyle=\ttfamily}
\begin{lstlisting}
(<@\bf{game}@> "Sudoku 4x4"  
 (<@\bf{{mode}@> 1)  
  
 (<@\bf{{equipment}@> { 
  (SudokuBoard 2) 
  (number P1 {1 2 3 4})
  })  
  
 (<@\bf{{rules}@> 
  (<@\bf{{start}@> { 
   (place 
    {4 1 3 3  1} 
    {1 5 7 13 15}
   )
  })
   
 (<@\bf{{play}@> (to {1 2 3 4} (empty)))
   
  (if (equal (count (empty)) 0)
   (<@\bf{end}@> 
    (if (and {
     (allDifferent (Row 0))
     (allDifferent (Row 1))
     (allDifferent (Row 2))
     (allDifferent (Row 3))   
     (allDifferent (Column 0))
     (allDifferent (Column 1))
     (allDifferent (Column 2))
     (allDifferent (Column 3)) 
     (allDifferent (set {0  1  4  5}))
     (allDifferent (set {2  3  6  7}))
     (allDifferent (set {8  9  12 13})) 
     (allDifferent (set {10 11 14 15}))   
     })
     (<@\bf{{result}@> 1 <@\bf{{Win}@>) 
     (<@\bf{result}@> 1 <@\bf{{Loss}@>) 
    )
   )
  )
 )
)
\end{lstlisting}
\caption{Game description of a Sudoku puzzle on a $4$$\times$$4$ grid in Ludii.}
 \label{fig:ludiiSudoku}
\end{figure}

\begin{figure}[!t]
\centering
\begin{tikzpicture}[scale=0.7]
\large
  \begin{scope}
    \draw (0, 0) grid (4, 4);
    \draw[very thick, scale=2] (0, 0) grid (2, 2);
    \setcounter{row}{1}
    \setrow { }{3}  {}{1}
    \setrow { }{}  {}{ }

    \setrow { }{1}  {}{3}
    \setrow { }{4}  {}{ }
  \end{scope}
  \end{tikzpicture}
\caption{Example of a Sudoku puzzle on a $4$$\times$$4$ grid.}
 \label{fig:Sudoku}
  \end{figure}

\subsection{XCSP}

XCSP3 \cite{BoussemartLP16} is a recent format to build integrated representations of combinatorial constrained problems. This new format is able to deal with mono/multi optimisation, many types of variables (integer, symbolic, graph, set, multi-set, etc.), cost functions, reification, views, annotations, variable quantification, distributed, probabilistic and qualitative reasoning. The new format is made compact, highly readable, and easy to parse. 
Similar to the philosophy of ludemes this format allows us to encapsulate the structure of the problem models, through the possibilities of declaring arrays of variables, and identifying syntactic and semantic groups of constraints. A competition for solvers using the XCSP format is organised annually,\footnote{XCSP competition: \url{http://www.xcsp.org/competition}} and as a result of this, most of the major constraint solvers developed by the CP community support XCSP. Accordingly, an XCSP instance can be solved using many different efficient solvers, including {\tt Abscon}, {\tt Choco}, {\tt Oscar} and {\tt SAT4J} to name of few. A website (\url{xcsp.org}) is developed conjointly with the format, which contains many models and series of instances, as well as different tools such as parsers in Java and C++. 

\subsection{From Ludii to XCSP}

\begin{figure}[!t]
\lstset{numbers=left, numberstyle=\tiny,stepnumber=1,numberfirstline=false, numbersep=5pt}
\scriptsize
\begin{xcsp}
<instance format="XCSP3" type="CSP">
<variables>
  <array id="x" size="[4][4]"> 1..4 </array>
</variables>
<constraints>

<instantiation class="hints"> 
    <list> x[0][1] x[1][1] x[1][3] x[3][1] x[3][3]</list>
    <values> 4 1 3 3 1 </values>
</instantiation>

<group>
  <allDifferent> %... </allDifferent>
    <args> x[0][] </args>
    <args> x[1][] </args>
    <args> x[2][] </args>
    <args> x[3][] </args>
    
    <args> x[][0] </args>
    <args> x[][1] </args>
    <args> x[][2] </args>
    <args> x[][3] </args>
    
    <args> x[0..1][0..1] </args>
    <args> x[0..1][2..3] </args>
    <args> x[2..3][0..1] </args>
    <args> x[2..3][2..3] </args>
</group>
</constraints>
</instance>
\end{xcsp}
\caption{Game description of a Sudoku puzzle on a $4$$\times$$4$ grid with XCSP.}
 \label{fig:XCSPLudii}
\end{figure}

\begin{figure}[!t]
\lstset{numbers=left, numberstyle=\tiny,stepnumber=1,numberfirstline=false, numbersep=5pt}
\scriptsize
\begin{xcsp}
<instantiation id='sol1' type='solution'>  
    <list> x[] </list>  
    <values> 
        3 4 1 2 2 1 4 3 1 2 3 4 4 3 2 1 
    </values>  
</instantiation>
\end{xcsp}
\caption{Solution of the XCSP instance in Figure \ref{fig:XCSPLudii}.}
 \label{fig:solSudoku}
\end{figure}

A Constraint Satisfaction Problem (CSP) consists of a set of variables -- each associated with a domain of possible values -- and a set of constraints that link the variables and define allowed combinations of values among them. 

Constraints can have several forms: two basic forms of them are (i) enumerating the list of allowed/forbidden tuples between variables and (ii) using common simple constraints in intention such as $=,\neq,<,\le,$ etc. Moreover, since the mid 90's, the concept of so-called {\it global constraints} has been introduced. Such constraints aim to improve the succinctness of constrained structures present in different problems, and are associated with more powerful filtering algorithms that can take into account the specificity of the formulated constraint to further reduce the domains of the variables, boosting the subsequent search for a solution. The first introduced and most famous global constraint is $\mathit{allDifferent}$, which states that each variable in its scope must take values different from all other ones. 

Interestingly, logic puzzles very often exhibit structure that can be encoded into global constraints. Furthermore, in Ludii, the library provides some ludemes similar to the main global constraints. For example, the global constraint $\mathit{allDifferent}$ is also used as a ludeme in Ludii (e.g. Figure \ref{fig:ludiiSudoku}), making the translation process easier.

Thanks to this proximity between the two languages, a translation process from Ludii to XCSP is possible in order to use the solution of the CSP problem generated as a sequence of moves for the Ludii game. Note that our study is restricted to one-player games ($Mode$ = $\{1\}$) using a single container ($|C^t| = 1$).

The variables and their domains are extracted from the $Equipment$ of a Ludii Game $\mathcal{G}$. Each CSP variable $v$ is generated from each grid cell of the container. The domain of each variable corresponds to the set $C^p$.

The generation of the constraints is obtained from the $Rules$ of $\mathcal{G}$. However, for the initial state ($start$ rules) and the terminal state ($end$ rules), the process is different. For puzzles, only the ludeme $place$ is used to put each component on different grid cells. This ludeme can easily be translated into constraints that assign the initial values to the corresponding variables. For the terminal state, each ludeme associated with a global constraint in the XCSP formalism is translated directly using the subset of variables corresponding to the region define on the ludeme. As an example, the translation of the $4$$\times$$4$ Sudoku described in Figure \ref{fig:ludiiSudoku} is given in Figure \ref{fig:XCSPLudii}. 

In this example, the lines $7$ to $10$ correspond to the $start$ rules in $\mathcal{G}$, and the lines between $12$ to $28$ correspond to the $end$ rules, each of them generated from the ludeme $\mathit{allDifferent}$ on different regions.

The solution provided by the CSP solver is translated into a sequence of Moves for the Ludii system. For each assignment, if the value associated is different to 0 (corresponding to an empty cell) and if in the $start$ rules no component is placed in the grid cell corresponding to the variable assigned, we apply a move to add the component corresponding to the value to the grid cell.

As an example, the CSP solution to the XCSP instance is given in Figure \ref{fig:solSudoku}. The corresponding sequence of moves is: {\tt Add(0,3)}, {\tt Add(2,1)}, {\tt Add(3,2)}, {\tt Add(4,2)}, {\tt Add(6,4)}, {\tt Add(8,1)}, {\tt Add(9,2)}, {\tt Add(10,3)}, {\tt Add(11,4)}, {\tt Add(12,4)}, {\tt Add(14,2)}.

\section{Experiments}

This section describes a number of experiments on different puzzles modelled with the ludemic approach.

{
\centering
\begin{table*}[!h]
\caption{Results to generate and solve puzzles with Ludii, XCSP and {\tt Abscon} (time in seconds)}
\label{table:results}
   \begin{center}
\scriptsize
%\begin{tabular}{|c|c|c}
\begin{tabular}{@{}lcccccr@{}}
\toprule
\bf Game & \bf Board Size & \bf {\tt \bf Ludii to XCSP in} & \bf {\tt \bf $\#$Variables} & \bf {\tt \bf Domain Size} & \bf {\tt \bf $\#$Constraints} & \bf {\tt \bf Solved in} \\
\midrule
%\rowcolor{vlgray} \multicolumn{2}{c}{Single player games} \\
%8 Puzzle  & 26,958 \\
%\rowcolor{vlgray} \multicolumn{2}{c}{Multi-player games} \\
\multirow{4}{*}{\bf Futoshiki} & $4$$\times$$4$ & 0.301 & 16 & 4 & 12 & 2.437 \\
 & $5$$\times$$5$ & 0.303 & 25 & 5 & 21 & 2.640 \\
 & $6$$\times$$6$ & 0.311 & 36 & 6 & 22 & 2.671 \\
 & $9$$\times$$9$ & 0.341 & 81 & 9 & 58 & 2.718 \\
\midrule
\multirow{3}{*}{\bf Latin Square} & $5$$\times$$5$ & 0.010 & 9 & 5 & 10 & 2.265 \\
 & $10$$\times$$10$ & 0.017 & 100 & 10 & 20 & 2.531 \\
 & $100$$\times$$100$ & 0.121 & 10,000 & 100 & 200 & 142.377 \\
\midrule
\multirow{3}{*}{\bf Magic Square} & $3$$\times$$3$ & 0.012 & 9 & 9 & 8 & 2.421 \\
 & $5$$\times$$5$ & 0.013 & 25 & 25 & 12 & 2.656 \\
 & $7$$\times$$7$ & 0.015 & 49 & 49 & 16 & 3.406 \\
\midrule
\multirow{2}{*}{\bf N Queens} & $4$$\times$$4$ & 0.011 & 16 & 2 & 61 & 3.125 \\
& $8$$\times$$8$ & 0.011 & 64 & 2 & 255 & 4.002 \\
% CP : Too long without symmetry breaking 
% & 16 $\times$ 16 & 0.011 & 256 & 2 & 1,021 & \textcolor{red}{todo} \\
% & 32 $\times$ 32 & 0.021 & 1,024 & 2 & 4,093 & \textcolor{red}{todo} \\
% & 64 $\times$ 64 & 0.034 & 4,096 & 2 & 16,356 & \textcolor{red}{todo} \\
% & 128 $\times$ 128 & 0.042 & 16,384 & 2 & 65,563 & \textcolor{red}{todo} \\
\midrule
\multirow{4}{*}{\bf Nonogram} & $5$$\times$$5$ & 0.013 & 25 & 2 & 10 & 1.328\\
 & $10$$\times$$10$ & 0.013 & 100 & 2 & 20 & 1.654\\
 & $20$$\times$$20$ & 0.014 & 400 & 2 & 40 & 1.843\\
 & $32$$\times$$32$ & 0.015 & 1,024 & 2 & 64 & 2.656\\
\midrule
\multirow{3}{*}{\bf Sudoku} & $9$$\times$$9$ & 0.010 & 81 & 9 & 27 & 2.421 \\
 & $16$$\times$$16$ & 0.012 & 256 & 16 & 48 & 2.734\\
 & $25$$\times$$25$ & 0.014 & 625 & 25 & 75 & 3.127\\
\bottomrule
\end{tabular}
\normalsize
\end{center} 
\end{table*}
}

\subsection{Setup}

In this section, we experiment with the translation and solving processes on some logic puzzles: Futoshiki, Latin Square, Magic Square, N Queens problems, Nonogram and Sudoku.
We provide the time used for each process and the size of the XCSP instance obtained by the translation process when given the number of variables, the size of the domain of each variable and the number of constraints. All experiments were conducted on a single core of an Intel(R) Core(TM) i7-8650U CPU @ 1.90 GHz, 2112 MHz with 16GB RAM.
To solve the XCSP instance, we use an open-source Java-written Constraint Solver called {\tt Abscon}, mainly developed by Christophe Lecoutre.
Ludii and {\tt Abscon} are running with the Java SE Development Kit 11. 

\subsection{Results}

First, as illustrated by Table \ref{table:results}, translations from Ludii to XCSP exhibit negligible runtimes, since they never exceed 1 second. We did not report them in a detailed manner due to lack of space, but we have tried larger sizes of puzzles and converting data towards XCSP formalism does not appear to be a barrier to this approach. 

For the solving part, we have deliberately tested games with different sizes, in order to get an idea of the practical limits of the approach. For example, converting and solving Latin Square (size 100) requires more than a minute, which can be seen as too long for some applications. However, all puzzles whose size makes them reasonably doable by humans are entirely solved within a few seconds : N Queens (sizes 4-8), Nonogram (sizes 5-32), Sudoku (sizes 9-25), etc. 

\subsection{Discussion}

These results allow us to apply online dynamic applications for all ``human-sized'' puzzles. For instance, Ludii can propose a series of (potentially randomly generated) puzzles and provide help, or prevent the user from mistakes, in a dynamic way through the use of CP solvers.

We intend to use Ludii to model the full range of Nikoli puzzles in a close future, and the translation proposed here is generic; once a new (NP-complete) puzzle is modelled in Ludii, it can be supported by a CP solver for the previous proposed applications (provide help and/or prevent mistakes), without requiring additional implementation work in most cases. 

It is important to keep in mind that we ``only'' propose to translate directly a puzzle from Ludii to an XCSP instance. Here no optimisation approaches such as heuristics or symmetries are used. This first approach only takes advantage of the global constraints. 

Consequently, a promising future work would be to include some data for the ludemes used to describe the puzzles in order to make it easier to apply some optimisations commonly used in CSP. As an example, the current direct translation for the Knight's Tour ($8$$\times$$8$) on XCSP is done in 2 seconds. However {\tt Abscon} requires 123 seconds to solve it. Handcrafting simple symmetries in the instance reduces the solving time to 19 seconds.

\section{Conclusion and Future Work}

This first work is a good example of cross-fertilisation. Indeed, while the Ludii system is now able to make use of decades of algorithmic progresses in Constraint Programming, just by stating puzzles, the XCSP ecosystem enriches its content of instances by providing new benchmarks for the CP community. 

In the future, the approach applied here on NP-complete puzzles could be extended to P-SPACE ones, such as Sokoban \cite{hearn02}. In addition, this work paves the way for a potential application of Constraint Programming on any multi-player games modelled on Ludii, in a similar spirit to WoodStock \cite{koriche17} -- a constraint-based approach for General Game Playing, and currently the best GGP agent, only compatible with GDL.

As another working track, a related work on \cite{Browne13}, proposes another approach called Deductive Search to solve logic puzzles. This approach is a breadth-first, depth-limited propagation scheme for the constraint-based solution of deduction puzzles, using simple logic operations found in standard constraint satisfaction solvers. This approach is particularly efficient for puzzles and has to be compared with the best CSP solvers in order to improve the resolution to any puzzle available on Ludii.

\section*{Acknowledgment.}

This research is part of the European Research Council-funded Digital Ludeme Project (ERC Consolidator Grant \#771292) run by Cameron Browne at Maastricht University's Department of Data Science and Knowledge Engineering. 

\bibliographystyle{IEEEtran}
\bibliography{IEEEabrv,References}

\end{document}